\documentclass[lettersize,journal]{IEEEtran}
\usepackage{amsmath,amsfonts}
\usepackage{algorithmic}
\usepackage{algorithm}
\usepackage{array}
\usepackage[caption=false,font=normalsize,labelfont=sf,textfont=sf]{subfig}
\usepackage{textcomp}
\usepackage{stfloats}
\usepackage{url}
\usepackage{verbatim}
\usepackage{graphicx}
\usepackage{cite}
\UseRawInputEncoding
\usepackage{color}
\usepackage{colortbl}
\usepackage{graphics} 
\usepackage{comment}
\usepackage{balance}
\usepackage{lastpage}
\usepackage{amsmath}
\usepackage{amssymb}
\usepackage{mathptmx}
\usepackage{bbm}
\usepackage{dsfont}
\usepackage{multirow}
\usepackage{bbding}
\usepackage{tikz}

\newcommand*{\circled}[1]{\lower.7ex\hbox{\tikz\draw (0pt, 0pt)%
    circle (.4em) node {\makebox[.5em][c]{\scriptsize #1}};}}

\newcommand{\etal}{\textit{et al. }}
\definecolor{lightpink}{RGB}{255,182,193}

\newcommand{\hh}[1]{{\color{black} #1}}
\newcommand{\xf}[1]{{\color{black} #1}}

\newcommand{\zh}[1]{{\color{black} #1}}
\newcommand{\huan}[1]{{\color{black} #1}}

\begin{document}

\title{Exploiting Low-level Representations for Ultra-Fast Road Segmentation}

\author{Huan Zhou, Feng Xue, Yucong Li, Shi Gong, Yiqun Li, Yu Zhou
\thanks{Corresponding author : Yu Zhou.}
\thanks{This research was supported by the National Natural Science Foundation of China (62176098, 61703049 and 51905185). 
}
\thanks{
H. Zhou, Y.C. Li, S. Gong, and Y. Zhou are with
Hubei Key Laboratory of Smart Internet Technology,
School of Electronic Information and Communications, 
Huazhong University of Science and Technology, Wuhan,
430074, China (e-mail:\{huanzhou, locoing, gongshi, yuzhou\}@hust.edu.cn).
}
\thanks{F. Xue is with the Department of Information Engineering and Computer Science, University of Trento, Trento, 38123, Italy (e-mail: xuefengbupt@gmail.com)}

\thanks{Y.Q. Li is with the School of Mechanical Science and Engineering, Huazhong University of Science and Technology, Wuhan, 430074, China (e-mail: liyiqun@hust.edu.cn).}
}


\maketitle

\begin{abstract}
\xf{Achieving real-time and accuracy on embedded platforms has always been the pursuit of road segmentation methods.
To this end, they have proposed many lightweight networks.
However, they ignore the fact that roads are ``stuff'' (background or environmental elements) rather than ``things'' (specific identifiable objects),
which inspires us to explore the feasibility of representing roads with low-level instead of high-level features.
Surprisingly, we find that the primary stage of mainstream network models is sufficient to represent most pixels of the road for segmentation.
Motivated by this,
we propose a \underline{L}ow-level \underline{F}eature \underline{D}ominated \underline{Road} \underline{Seg}mentation network (LFD-RoadSeg).}
Specifically,
LFD-RoadSeg employs a bilateral structure.
The spatial detail branch is firstly designed to extract low-level feature representation for the road by the first stage of ResNet-18.
To suppress texture-less regions mistaken as the road in the low-level feature,
the context semantic branch is then designed to extract the context feature in a fast manner.
To this end,
in the second branch,
we asymmetrically downsample the input image and design an aggregation module to achieve comparable receptive fields to the third stage of ResNet-18 but with less time consumption.
Finally,
to segment the road from the low-level feature,
a selective fusion module is proposed to calculate pixel-wise attention between the low-level representation and context feature,
and suppress the non-road low-level response by this attention.
\xf{On KITTI-Road, LFD-RoadSeg achieves a maximum F1-measure (MaxF) of 95.21\% and an average precision of 93.71\%, while reaching 238 FPS on a single TITAN Xp and 54 FPS on a Jetson TX2, all with a compact model size of just 936k parameters.}
The source code is available at https://github.com/zhouhuan-hust/LFD-RoadSeg.
\end{abstract}


\begin{figure}[!t]
\centering
\includegraphics[width=1.0\linewidth]{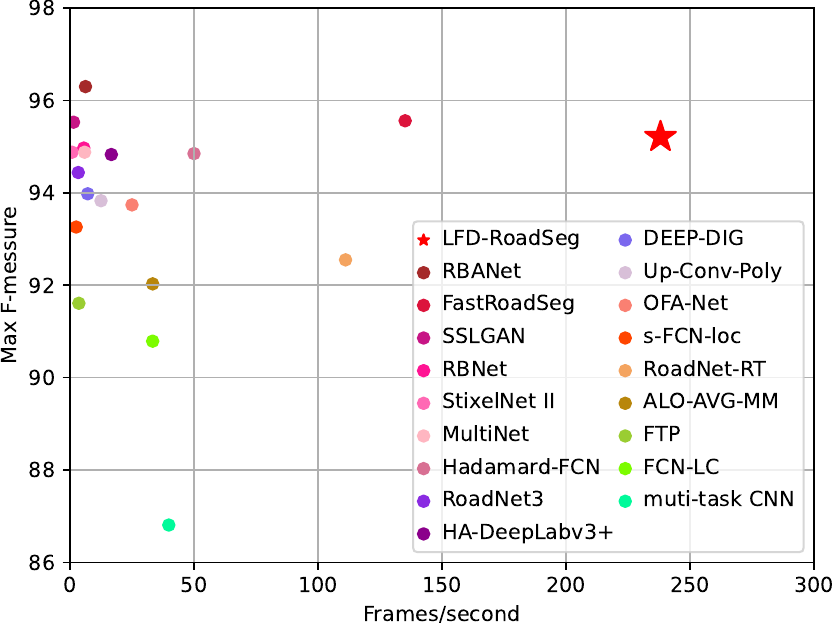}
\caption{Accuracy (MaxF) vs. efficiency (FPS) for various monocular road segmentation algorithms on KITTI-Road benchmark.}
\label{FPS}
\end{figure}

\section{Introduction}
\IEEEPARstart{V}{isual} road segmentation has become the fundamental scene understanding approach for autonomous driving and robots \cite{xue2021boundary, xue2019novel, xue, ZheLiu-AAAI2019-TANet, lu2019occlusion, 9605219, xue_indoor_2023, kherraki2023efficient, xu2023comparative, zhou2023occlusion}.
Although it was originally introduced more than 15 years ago,
the improvement of embedded platforms and deep networks has enabled the deployment of road segmentation on autonomous driving systems only in recent years.
\xf{Due to scarce computing resources,
embedded platforms require models to be low-latency and lightweight,
which is the goal that the recent lightweight road segmentation networks have been pursuing.
The efficiency of these methods is shown in Fig. \ref{FPS}.}
Oliveira \etal \cite{Oliveira:1} proposed a lightweight FCN-like network that achieves 12 FPS on an NVIDIA TITAN X GPU.
Oeljeklaus \etal \cite{Oeljeklaus:1} appended two decoders of object detection and road segmentation after the inception-v2 network to realize a fast multi-task CNN,
achieving a speed of 187.9 milliseconds (ms) per image on an NVIDIA Jetson TX2.
Bai \etal \cite{bai2020roadnet} designed a lightweight segmentation network with a bilateral structure, namely, RoadNet-RT,
achieving a speed of 9 ms per image on a GTX 1080 GPU.
Gong \etal \cite{gong2022fastroadseg} proposed a fast encoder-decoder network that further increases the speed to 135 FPS on a TITAN Xp GPU while achieving MaxF over 95\%.
\xf{Overall, it is not too much to be faster for the road segmentation models on embedded systems.}

\begin{figure}
\centering
\includegraphics[width=0.85\linewidth]{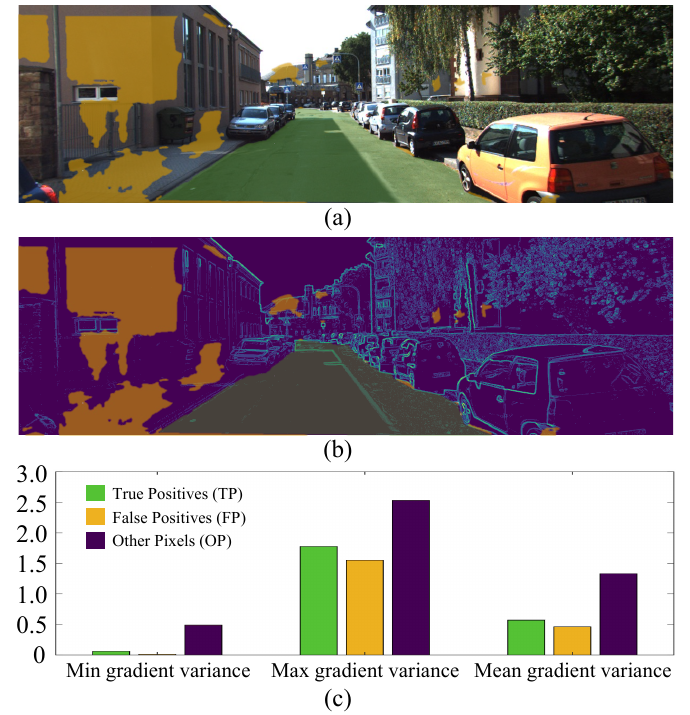}
\caption{(a) shows an example image, where yellow indicates FP, and green for TP.
(b) shows the gradient of this image.
(c) shows the minimal, maximal, and mean gradient variance of TP, FP and OP in the validation set of the KITTI-Road dataset.
}
\label{fig:var}
\end{figure}

Although these approaches vary in network topology and training process,
\xf{they overlook a crucial characteristic of the road:
\textit{roads are `stuff', namely background or environmental elements in an image, rather than `things', which refer to specific identifiable objects}.
Therefore, the classification of road pixels depends much less on semantic information than that of objects with semantic categories.}
To make a deep exploration,
we implement four networks that respectively utilize the 1$^{st}$, 2$^{nd}$, 3$^{rd}$ and 4$^{th}$ stages' feature maps of ResNet-18 to segment the road,
and their performances are shown in Table \ref{tab:intro}.
Apparently,
the model using the 1$^{st}$ stage feature is obviously inferior in precision,
while it obtains a recall rate as high as 93.58\% and saves 94.33\% parameters compared to the model using the 3$^{rd}$ stage.
This phenomenon demonstrates that the model using the 1$^{st}$ stage feature finds most pixels of the road, but it suffers from false detection of several areas similar to the road surface,
which is consistent with the example result in Fig. \ref{fig:var} (a).
Subsequently,
we delve deeper into the characteristics of these mis-detected pixels.
To this end,
we group all pixels in the prediction into three sets,
i.e., the true positives (TP),
the false positives (FP),
and other pixels (OP).
Then, the RGB gradient variance for each set is computed as an indicator of the texture intensity.
Fig. \ref{fig:var} (c) shows the gradient variance of TP, FP, and OP.
Intuitively,
TP and FP typically have lower texture intensity than OP,
and FP has an even lower texture intensity than TP,
meaning that FP is generally situated in areas with weak texture.
With all the above in mind,
through extracting the context for correlating each weak texture area in a fast manner,
the false detection can be eliminated effectively,
while the advantages of minimal parameter and time cost can also be retained.

\begin{table}
\caption{Comparison of the models using different stages of ResNet-18 for road segmentation.
The \huan{highlighted} row compares the 1$^{st}$ and 3$^{rd}$ stages,
and the \huan{bolded values show} the best results. The input image resolution is $375 \times 1240$.}
\label{tab:intro}
\centering
\begin{tabular}{|l|p{0.6cm}<{\centering}p{0.6cm}<{\centering}|p{0.6cm}<{\centering}p{0.6cm}<{\centering}|p{1cm}<{\centering}p{1cm}<{\centering}|}
\hline
Feat used & MaxF & AP & PRE & REC & Params & Time(ms)\\
 \hline
 \hline
1$^{st}$ Stage & 90.86 & 86.30 & 88.29 & 93.58 & \textbf{157,634} & \textbf{1.80} \\
\rowcolor{lightpink}1$^{st}$ vs. 3$^{rd}$ & -5.35 & -7.81 & -8.27 & -3.28 & -94.33\% & -45.29\% \\
2$^{nd}$ Stage & 95.53 & 93.10 & 95.25 & 95.81 & 683,330 & 2.61 \\
3$^{rd}$ Stage & \textbf{96.21} & \textbf{94.11} & \textbf{96.56} & \textbf{95.86} & 2,783,298 & 3.29 \\
4$^{th}$ Stage & 95.08 & 93.74 & 94.87 & 95.29 & 11,177,538 & 4.58 \\
\hline
\end{tabular}
\end{table}

\xf{Based on the insights from the experiment above,
we propose a low-level feature dominated road segmentation network (LFD-RoadSeg) that follows the bilateral structure.
For the spatial detail branch,
we employ the primary stage of lightweight backbone networks to extract the low-level road representation,
ensuring high resolution and low latency.
To eliminate non-road response in the low-level features,
we design a context semantic branch in a fast manner to capture context as a supplement.
For this context semantic branch,
we first propose asymmetric downsampling to enable contextual features to have a large horizontal receptive that has been proven to be crucial for street scenes.
Then, we design a lightweight aggregation module to capture the context with comparable receptive fields to the ResNet-18’s 3$^{rd}$ stage that is proven effective in road segmentation (Table \ref{tab:intro}). 
Finally,}
\huan{
we design a selective fusion module to segment road regions from low-level road representations.
This module leverages context-based spatial attention to suppress non-road responses in low-level features.
\xf{The KITTI-Road, Cityscapes and CamVid datasets are employed to evaluate our method.
In the experiments, LFD-RoadSeg achieves excellent effectiveness and the fastest speed so far,}
which can be observed in Fig. \ref{FPS}.
}

In summary, the contributions of this paper are as follows:
\begin{itemize}
\item \huan{
\xf{We reveal the ``stuff'' characteristic of roads overlooked by the previous road segmentation methods,
and find that the primary stage of mainstream networks is adequate to extract road features.}
This motivates us to represent the road by low-level features.
}

\item \xf{We propose LFD-RoadSeg.
It novelly leverages low-level road representation as the basis for segmentation and employs the proposed asymmetric downsampling and aggregation modules to accelerate context extraction.}

\item \xf{LFD-RoadSeg boosts speed to 238 FPS (twice the previous fastest method) on a single TITAN Xp and 54 FPS on a Jetson TX2 with only 936k parameters,
but still gains a decent MaxF of 95.21\% on KITTI-Road.
Thus, our method advances the practicability of road segmentation.}

\end{itemize}




\section{RELATED WORK}
\label{related}

In this section,
we briefly review the monocular road segmentation and the bilateral network for semantic segmentation,
which are closely related to our approach.

\subsection{Monocular Road Segmentation}
In the early years of researching monocular road segmentation,
the community \cite{Alvarez:1,Alvarez:2,Chacra:1,Sturgess:1, ma2017object, zhou2016similarity, NIPS2012_3e313b9b, 2014ONLINE} mainly focused on designing low-level features,
such as color, edge, texture, etc., to represent and classify the road at pixel level or patch level.
Later, several works tried to introduce global information to improve the reliability of road representation.
Vitor \etal \cite{Vitor:1} \zh{designed} the global probabilistic model to aggregate multiple descriptors to represent the road.
Mario \etal \cite{Mario:1} \zh{employed} conditional random fields (CRF) to model dependencies across the whole image.
Although these approaches usually achieve a road region with crisp boundary,
they perform poorly in complex scenes that contain illumination change or tree shade,
due to the poor generalization of the handcrafted feature.
In addition,
these works are not GPU-accelerated generally and thus far from real-world applications.

In recent years,
monocular road segmentation has been greatly boosted by the development of neural networks.
Several early approaches \cite{Alvarez,Mendes:1} attempted to classify pixels or patches of the road by using neural networks as the classifier.
The later methods \cite{Oliveira:1,rbanet,teichmann2018multinet,gong2022fastroadseg} \zh{followed} the pipeline of deep semantic segmentation \cite{deeplab,fcn,pspnet}.
Teichmann \etal \cite{teichmann2018multinet} designed a unified architecture to conduct road segmentation, object detection, and scene classification simultaneously.
Sun \etal \cite{rbanet} \zh{focused} on the attention mechanism to recover detailed information around the road boundary.
Compared to the handcrafted methods,
these \huan{methods achieve} superior generalization,
which makes them perform accurately in many road scenes.
However,
they still have a non-negligible computational burden on a mobile platform.

\huan{
\subsection{Methodologies for Embedded Systems}
With the increasing application of road segmentation in the autonomous driving system,
road segmentation for embedded systems has received increasing attention.
Oeljeklaus \etal \cite{Oeljeklaus:1} proposed a fast multitask CNN for perceiving objects and the road,
which achieves a speed of \xf{5.32 FPS} ($375\times1240$) on the Jetson TX2 \cite{tx2} embedded platform.
Bai \etal \cite{bai2020roadnet} designed a road segmentation network RoadNet-RT optimized for FPGA,
which achieves a speed of \xf{111 FPS} ($280\times960$) on a GTX 1080 GPU. With the same setting of the experiment, Gong \etal \cite{gong2022fastroadseg} proposed a fast encoder-decoder network to speed up road segmentation to 31 FPS ($187\times620$) on the Jetson TX2 embedded platform, while achieving a MaxF above 95\%, much higher than the works mentioned above.
\xf{However, they neglect that roads are ``stuff'',
meaning that road segmentation relies more on low-level features.
This inspires us to propose LFD-RoadSeg which achieves faster speed,
lighter weight and better trade-off than the previous methods.}
}

\begin{figure*}[!t]
\centering
\includegraphics[width=1.0\linewidth]{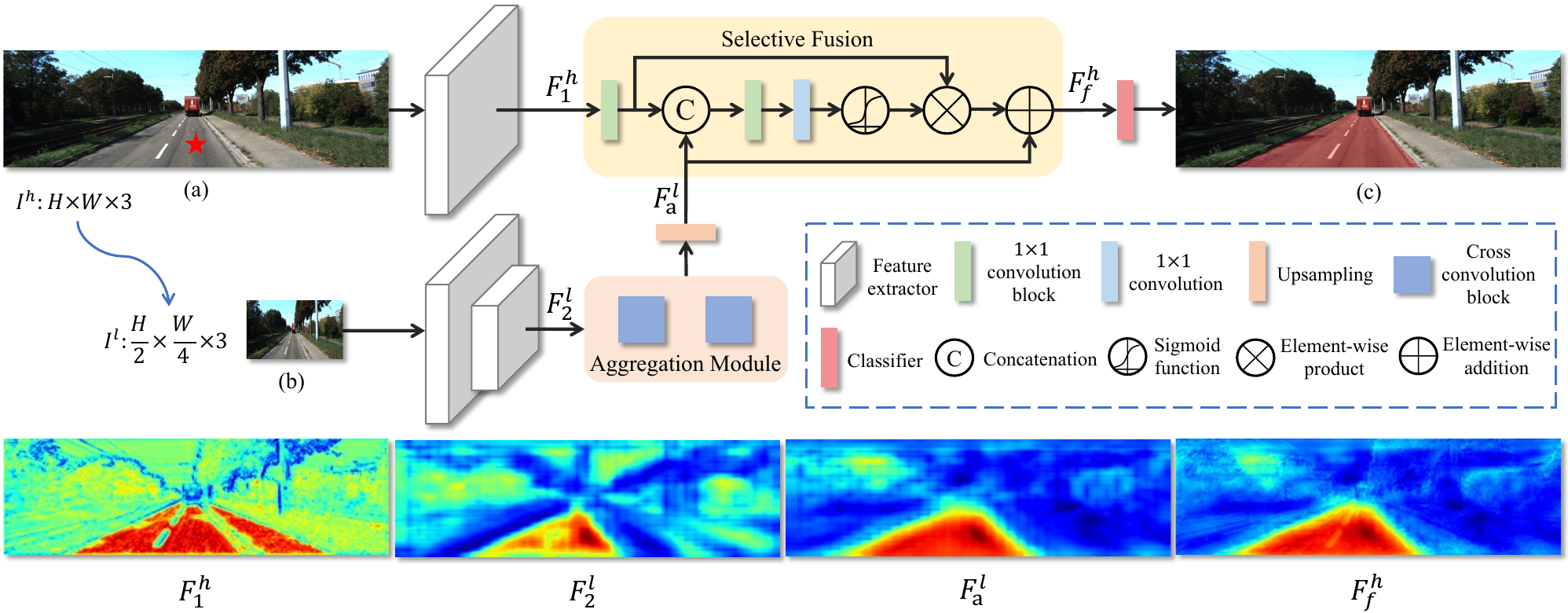}
\caption{The network architecture of LFD-RoadSeg. (a) is the input image ${I}^{h} \in \mathbb{R}^{H \times W \times 3}$ of spatial detail branch. (b) is the input image ${I}^{l} \in \mathbb{R}^{\frac{H}{2} \times \frac{W}{4} \times 3}$ of context semantic branch. (c) is the visualization of the road prediction result, where the area covered by red is the road.
The red star in the input image refers to the query point. The feature cosine similarity heat maps represent the correlation of the features between the query point and all others.}
\label{overview}
\end{figure*}

\subsection{Bilateral Network for Semantic Segmentation}
The bilateral networks \cite{yu2018bisenet} extract the spatial details and categorical semantics separately.
And it is popular in the community of semantic segmentation due to its faster speed than other structures.
Dong \etal \cite{Dong2021} designed a lightweight baseline network with atrous convolution and a distinctive atrous spatial pyramid pooling for semantic extraction.
BiSeNetv2 \cite{yu2021bisenet} proposed a bilateral guided aggregation layer to enhance the mutual connections of the two branches.
CABiNet \cite{kumaar2021cabinet} designed a context branch with lightweight versions of global aggregation and local distribution blocks.
Different from the above methods,
which use two independent branches,
Fast-SCNN \cite{poudel2019fast}, EACNet \cite{li2021eacnet} and DDRNet \cite{hong2021ddrnet} all utilized a shared trunk and two parallel branches with different resolutions.
ContextNet \cite{poudel2018contextnet} reduced the input image's resolution for the semantic branch to a quarter of the original image to accelerate the inference process.


\huan{
\subsection{Motivation}
\xf{LFD-RoadSeg is motivated by the fact that roads are ``stuff''.
According to \cite{Kirillov_2019_CVPR}, ``stuff'' is the background and environmental elements in the image, rather than ``things'' (such as person and car) that rely on semantic features in classification.
Thus, the same ``stuff'' area has a similar texture,
leading us to believe that road pixels can be classified by using low-level features.
To this end, we compare the differences of various stages in road segmentation.
Consequently,
we find that the first stage of ResNet is adequate for representing road,
as evidenced by the high recall rate of 93.58\% in Table \ref{tab:intro} and the green area in Fig. \ref{fig:var}.
However, the first stage of ResNet suffers from the false positives, 
which is indicated by the low precision and yellow area in Fig. \ref{fig:var}.
This inspires us to design a lightweight context semantic branch to quickly extract correlation between each area,
which gives the indicators for reducing the non-road response in the low-level feature.}
}


\section{Approach}
\label{method}
In this section, 
we first describe the specific network architecture in Sec. \ref{sec:network} in detail, including the spatial detail branch, the context semantic branch and the selective fusion module. Then the loss function dedicated to hard negative sample mining we used is introduced in Sec. \ref{sec:loss}.


\subsection{The Proposed Network Architecture}
\label{sec:network}
Inspired by the comparisons in Table \ref{tab:intro},
we represent the road by low-level features to achieve ultra-fast road segmentation.
As illustrated in Fig. \ref{overview},
the overall structure of our network includes two branches and a feature selective fusion module.
And \huan{regarding} the two branches,
one is a high-resolution spatial detail branch,
and the other is a low-resolution context semantic branch.

\textit{1) \textbf{Spatial Detail Branch:}}
The spatial detail branch is to capture most pixels of the road from the input RGB image ${I}^{h} \in \mathbb{R}^{H \times W \times 3}$ with only a small amount of convolutions.
\xf{Although our method can be applied to various backbone networks,
for representation simplification,
we describe the following network structure based on ResNet-18 by default.}
To be specific,
let $F^{h}_i$ be the output feature maps of the $i^{th}$ stage of a ResNet-18 network when taking ${I}^{h}$ as input.
And the spatial detail branch employs the 1$^{st}$ stage of ResNet-18 as the backbone and outputs the feature $F_{1}^{h} \in \mathbb{R}^{\frac{H}{4} \times \frac{W}{4} \times 64}$.
According to Table \ref{tab:intro},
the extremely light structure of this branch encodes most of the road pixels and preserves high resolution with low computational cost.

\textit{2) \textbf{Context Semantic Branch:}}
The context semantic branch aims to capture contextual information in a fast manner for suppressing the texture-less non-road region in $F_{1}^{h}$.
Specifically,
this branch employs two designs to achieve an extremely fast speed of context extraction.
\huan{Firstly,
since the correlation between horizontal areas is crucial for road segmentation \cite{gong2022fastroadseg},
the input RGB image ${I}^{h}$ is downsampled by a factor of 4 horizontally and by a factor of 2 vertically to obtain a low-resolution RGB image ${I}^{l} \in \mathbb{R}^{\frac{H}{2} \times \frac{W}{4} \times 3}$.
The resizing operation achieves a larger horizontal receptive field and a decent inference time.}
Secondly,
since the 3$^{rd}$ stage of ResNet-18 outperforms others in Table \ref{tab:intro},
the context semantic branch requires the same receptive field as the 3$^{rd}$ stage.
To this end,
this branch utilizes the first two stages of ResNet-18 to obtain the feature $F_{2}^{l} \in \mathbb{R}^{\frac{W}{16} \times \frac{W}{32} \times 128}$ from ${I}^{l}$,
and then appends a newly designed aggregation module to achieve a similar receptive field as the 3$^{rd}$ stage,
but at a faster speed than the first three stages of ResNet-18.

Next, we elaborate on the structure of the newly designed aggregation module and discuss the advantages of the above design in terms of computation and parameter amount.


\begin{figure}[!t]
\centering
\includegraphics[width=1.0\linewidth]{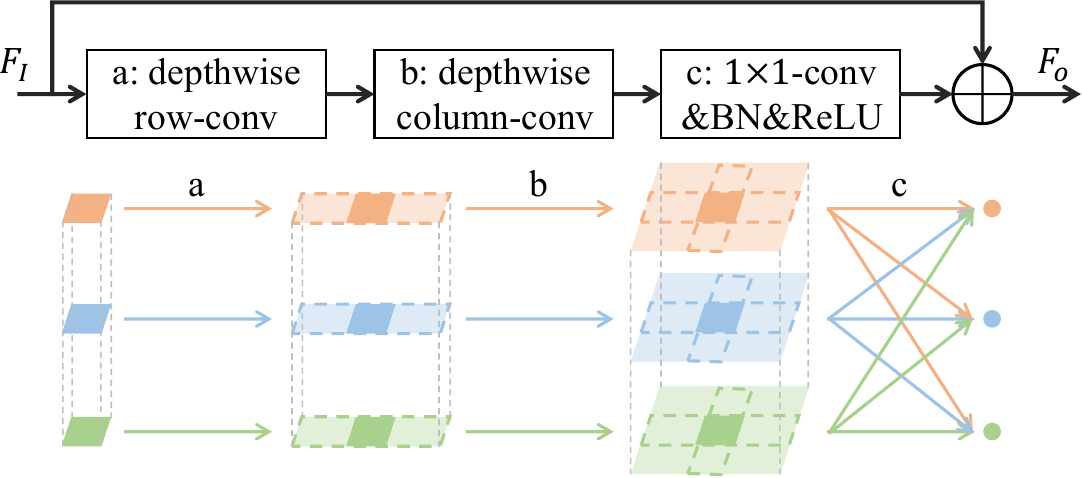}
\caption{Structure of the cross convolution block. $\bigoplus$ denotes the element-wise sum operation.}
\label{cross}
\end{figure}

\textbf{Aggregation Module.}
\hh{The 3$^{rd}$ stage of the original ResNet-18 has parameters up to 2.1M (see Table \ref{tab:intro}, 2,783,298 - 683,330 = 2,099,968), and the aggregation module implements similar feature extraction capabilities with a lighter structure.}
It consists of two cross convolution blocks,
and the structure of each cross convolution block is shown in Fig. \ref{cross}.
\hh{In each cross convolution block,
assuming that $F_I$ denotes the input feature,
we use a $1 \times 5$ row convolution and a $5 \times 1$ column convolution to simulate a large-kernel convolution for context feature extraction on $F_I$.}
Both the row and column convolution are depth-separable and aggregate context information in each channel respectively.
Then, a \huan{${1\times1}$} convolution is employed to fuse all channels.
Note that, to gain a larger horizontal receptive field,
the row convolution in the first cross convolution block has a dilation of 2.
Finally, we add the fused feature and the input feature $F_I$ element-wisely to reserve the detail contained in the input feature,
and the output feature $F_O$ can be formulated as:
\begin{equation}
\begin{split}
\huan{F_O = {B}_{1\times1}({C}({R}(F_I))) + F_I}
\end{split}
\label{conv}
\end{equation}
where \huan{${R}$} denotes the depthwise row convolution (row-conv for short), \huan{${C}$} denotes the depthwise column convolution (column-conv for short), and \huan{${B}_{1\times1}$ denotes the $1\times1$ convolution block which contains a \huan{${1\times1}$} convolution (\huan{${1\times1}$}-conv for short), a batch normalization and a ReLU activation.}
By using two cross convolution blocks connected in series,
we extract the context feature from the feature $F_{2}^{l}$,
which is followed by \huan{an} upsampling operation to align the feature resolution to $F_1^h$, namely $\frac{H}{4} \times \frac{W}{4}$.
And we obtain a high-resolution context feature $F_a^l \in \mathbb{R}^{\frac{W}{4} \times \frac{W}{4} \times 128}$.


\begin{table}[!t]
\caption{Model complexity and inference time comparison.
The input image resolution of the first row is $375 \times 1240$,
and the resolution of the second row is $187 \times 310$.}
\label{csb}
\centering
\begin{tabular}{c|c|c|c}
\hline

  & MACs[G] & Params[M] &Time(ms) \\
 \hline
 \hline
 
Three stages of ResNet-18 &13.23  &2.78  &3.24\\
Context semantic branch &1.21\color{red}{(-91\%)}  &0.72\color{red}{(-74\%)} &1.90\color{red}{(-41\%)}\\
\hline
\end{tabular}
\end{table}

\begin{figure}[!t]
\centering
\includegraphics[width=1.0\linewidth]{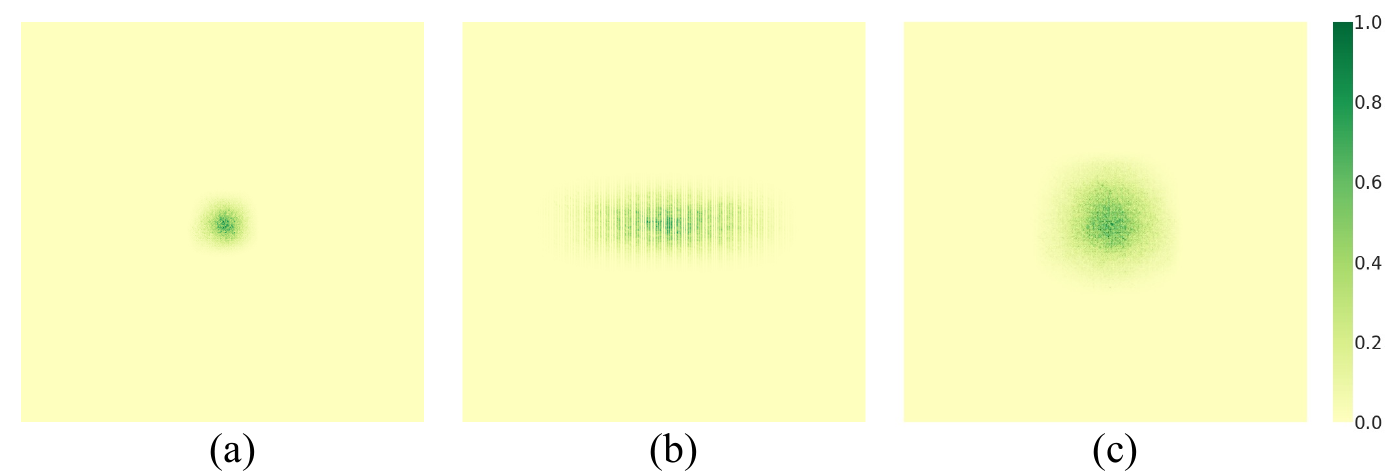}
\caption{Visualization of the model effective receptive field. (a) is the effective receptive field for the first three stages of ResNet-18. (b) is the effective receptive field for the context semantic branch. (c) is the effective receptive field for all four stages of ResNet-18.}
\label{receptive}
\end{figure}

\textbf{Discussion.} 
In the section above,
two designs are given to make the context semantic branch have similar discriminative power as the first three stages of ResNet-18 but with fewer parameters and computations.
\hh{
One is to reduce the input image size,
and the other is the aggregation module that replaces} ResNet-18's 3$^{rd}$ stage.
Compared to the original first three stages of ResNet-18,
the context semantic branch reduces the MACs (multiply-accumulate operations) by 91\%,
parameters by 74\%,
and inference time by 41\%, as shown in Table \ref{csb}.
Furthermore, we visualize the effective receptive field of ResNet-18 and the context semantic branch in Fig. \ref{receptive}.
It can be seen that the context semantic branch has a comparable receptive field to ResNet-18's 3$^{rd}$ stage in the vertical direction.
Note that,
due to the asymmetric downsampling and the dilation row-conv,
our method has a larger receptive field in the horizontal direction,
which is good for extracting correlation between different areas (e.g., drivable road, sidewalk, rail track, etc.) that are in the horizontal direction of each other \cite{gong2022fastroadseg}.

\textit{3) \textbf{Selective Fusion:}}
The high-resolution road representation $F_{1}^{h}$ obtained by the spatial detail branch suffers from weakly textured non-road regions that are easily misclassified.
In this section,
we propose a selective fusion module to remove these non-road regions by the context feature $F_{a}^{l}$ obtained from the context semantic branch.
Specifically,
the structure of the selective fusion module is shown in Fig. \ref{overview}.
Firstly, we use a 1$\times$1 convolution block to adjust the number of channels of $F_1^h$ to be the same as that of $F_a^l$, namely 128.
Then we concatenate the output feature of both branches, namely $F_1^h$ and $F_a^l$,
and calculate the pixel-wise attention between them by a 1$\times$1 convolution block, a \huan{${1\times1}$} convolution and a sigmoid function,
which can be formulated as:
\begin{equation}
\begin{split}
\huan{F_{attention} = {S}({P}({B}_{1\times1}({B}_{1\times1}(F_1^h) \circled{C} F_a^l)))}
\end{split}
\label{conv}
\end{equation}
where $F_{attention} \in \mathbb{R}^{\frac{H}{4} \times \frac{W}{4} \times 1}$ is the spatial attention weight.
\huan{${S}$} denotes sigmoid function,
\huan{${P}$} denotes \huan{${1\times1}$} convolution,
and $\circled{C}$ denotes concatenation.
Finally, the spatial attention weight $F_{attention}$ is employed to select the road area from the low-level representation $F_1^h$,
and the context feature also serves as the complementary expression of the road:
\begin{equation}
\begin{split}
\huan{F_f^h = F_{attention} \otimes {B}_{1\times1}(F_1^h) \oplus F_a^l}
\end{split}
\label{conv}
\end{equation}
where $\otimes$ denotes the element-wise product operation,
$\oplus$ denotes the element-wise sum operation,
and the attention weights $F_{attention}$ are applied to all channels of the high-resolution branch feature.
Finally, the fused feature is fed into a pixel-wise classifier, namely a block containing a \huan{\huan{${1\times1}$}} convolution,
a batch normalization,
a ReLU activation and a \huan{\huan{${1\times1}$}} convolution, to segment the road.
As shown in Fig. \ref{overview},
after the selective fusion,
the feature response of the sidewalk on the right side of the road in $F_1^h$ is suppressed, and the road features are enhanced.

\textbf{Differences between RoadNet-RT and LFD-RoadSeg:}
Although the overall structure of LFD-RoadSeg is roughly similar to RoadNet-RT \cite{bai2020roadnet},
there are still three differences between them.
Firstly,
we determine the network depth and receptive field of the two branches by experiments, rather than solely relying on intuition.
Secondly,
we propose the aggregation module to more effectively capture the contextual information, instead of using ASPP and global attention.
Thirdly,
during fusing the two branches' features,
we calculate the spatial attention to express the relation between the low-level feature and the context feature,
and utilize it to suppress texture-less non-road regions in low-level features.
Therefore,
our fusion process is different from RoadNet-RT which performs channel attention fusion on the two branches.


\subsection{Loss Function}
\label{sec:loss}
Since road segmentation is a binary classification problem, binary cross-entropy loss is generally used as the loss function of the classifier to supervise the final output. Let $N$ \huan{denote} the number of pixels, $i$ and $j$ are the pixel index in the image, $y \in \{0,1\}$ denotes the ground truth and $p$ is the predicted confidence,
the binary cross-entropy loss is formulated as:
\begin{equation}
\begin{split}
\bold{L}_{\rm{bce}} = -\frac{1}{N}\sum_{i,j}y^{i,j}log(p^{i,j})+
(1-y^{i,j})log(1-p^{i,j})
\end{split}
\label{bceloss}
\end{equation}

As we know, the texture-less road and some texture-less areas such as vegetation, sky and buildings are easier to distinguish than highly textured areas.
And the drivable road boundaries, sidewalks, and abnormal road areas such as overexposure and shadow are prone to be misclassified.
Therefore, in the training process,
we utilize a strategy similar to OHEM \cite{ohem} to mine difficult-to-segment pixels.
Given a confidence threshold  $\lambda_{b}$ in a batch,
we only perform gradient backpropagation for pixels whose predicted confidence $p$ is less than the threshold $\lambda_{b}$.
Let $\mathds{1}(\cdot)$ denote the indicator function, and
the entire loss function of the network $\bold{L}_{\rm{main}}$ is formulated as:
\begin{equation}
\begin{split}
\bold{L}_{\rm{main}} = \frac{1}{N}\sum_{i,j}\mathds{1}(p^{i,j}<\lambda_{b})\bold{L}_{\rm{bce}}^{i,j}
\end{split}
\label{ohem}
\end{equation}

\section{experiments}
\label{experiment}
In this section, we first describe the datasets and evaluation metrics in Sec. \ref{sec:dataset}. Then the detail of the network training is given in Sec. \ref{sec:train}. Sec. \ref{sec:result} reports the quantitative and qualitative results of our method. Finally, the ablation study of each component and the discussion on the input image size of the context semantic branch are given in Sec. \ref{sec:ablation} and Sec. \ref{sec:discussion}, respectively. 

\begin{table*}[!t]
\scriptsize
\caption{Comparison with prior representative road segmentation works on KITTI-Road dataset \\(\huan{``-" means it is not mentioned in the official database and the original paper})}
\label{leaderboard}
\centering
\begin{tabular}{c | c | c | c c c c c c | c | c}

\hline
Method  &Sensor &Input shape &  MaxF(\%)$\uparrow$  & AP(\%)$\uparrow$ & PRE(\%)$\uparrow$ & REC(\%)$\uparrow$ & FPR(\%)$\downarrow$ & FNR(\%)$\downarrow$ & Time(ms)$\downarrow$ & Device \\
\hline
\hline
RBANet \cite{rbanet}    &Cam. &$360\times720$ &    {96.30}    &    89.72    &    95.14    &    {97.50}    &    2.75    &    {2.50}    &    160  &TITAN Xp \\

FastRoadSeg  \cite{gong2022fastroadseg}    & Cam. &$375\times1240$ &95.56    &    93.89    &    95.53    &    95.59    &    {2.47}    &    4.41    & 7.4  &TITAN Xp \\
SSLGAN  \cite{han2018semisupervised}  &Cam.& $375\times1242$   &{95.53}    &90.35    &    {95.84}    &    95.24    &    {2.28}    &    4.76    &    700 &TITAN X\\

RBNet \cite{rbnet}       &Cam. &$300\times900$ &    94.97    &    91.49    &    94.94    &    95.01    &    2.79    &    4.99    &   180 &Tesla K20c \\

StixelNet II \cite{garnett2017real} &Cam.&$370\times800$  &   94.88    &    87.75    &    92.97    &    96.87    &    4.04    &    3.13    &    1200  &Quadro M6000\\
MultiNet  \cite{teichmann2018multinet}   &Cam. &$384\times1248$ &    94.88    & {93.71}    &    94.84    &    94.91    &    2.85    &    5.09    &170 &GTX 1080 \\
Hadamard-FCN  \cite{oeljeklaus2021integrated}   &Cam. &$375\times1242$ &    94.85    &    91.48    &94.81    &    94.89    &    2.86    &    5.11    & 20 &TITAN X\\
HA-DeepLabv3+  \cite{fan2021learning}   &Cam. &- &    94.83    &    93.24   &    94.77    &    94.89    &    2.88    &    5.11    &60  & -\\
RoadNet3  \cite{roadnet3}   &Cam.  &$160\times600$ &    94.44    &    93.45    &    94.69    &    94.18    &    2.91    &    5.82    &    300 &GTX 950M\\

DEEP-DIG \cite{munoz2017deep}   & Cam. &- &    93.98   &    93.65   &    94.26   &    93.69   &    3.14   &    6.31   &    140 & TITAN X\\
Up-Conv-Poly \cite{up-conv-poly} &Cam. &$500\times500$ &   93.83   &    90.47   &    94.00   &    93.67   &    3.29   &    6.33   &80 & TITAN X\\
OFA-Net \cite{zhang2019one}      &Cam.&- &   93.74   &    85.37   &    90.36   &    97.38   &    5.72   &    2.62   &40  &- \\
s-FCN-loc \cite{wang:3}   &Cam.   &$500\times500$ & 93.26   & -        & 94.16     & 92.39      & 3.16         & 7.61  & 400 &Tesla K80 \\

RoadNet-RT \cite{bai2020roadnet}  &Cam.  &$280\times960$ &  92.55   &    93.21   &    92.94   &    92.16   &    3.86   &    7.84   & 9  & GTX 1080 \\
ALO-AVG-MM \cite{Reis:1}    &Cam. &$192\times624$ &  92.03   &    85.64   &    90.65   &    93.45   &    5.31   &    6.55   & 30  & GTX 1080\\   
\hline
\textbf{LFD-RoadSeg}      &Cam.  &$375\times1240$ &    \textbf{95.21}    &   \textbf{93.71}    &\textbf{95.35}    &   \textbf{95.08}   &    \textbf{2.56}    &   \textbf{4.92}    &    \textbf{4.2}  &TITAN Xp\\

\hline
\end{tabular}
\end{table*}

\subsection{Dataset and Evaluation}
\label{sec:dataset}
\textit{1) Dataset:}
Three datasets are employed to evaluate the effectiveness of our method.
KITTI-Road \cite{kitti} is a \huan{real-world} road segmentation dataset containing 289 training images and 290 testing images. It has three categories of road scenes, Urban Unmarked (UU), Urban Marked (UM) and Urban Multiple Marked (UMM). URBAN is a combination of the three above.
The resolution of KITTI-Road training images ranges from 370 $\times$ 1224 to 375 $\times$ 1242. For the convenience of training, we unify them as 375 $\times$ 1240 \huan{by padding operation}. The evaluation is done by the official online evaluation server.
Following \cite{gong2022fastroadseg}, in the ablation experiments, we use 5-fold cross-validation on the training images, and the experimental results are expressed as (mean $\pm$ standard deviation).
Cityscapes \cite{Cityscapes} is a real-world dataset of urban street 
scenarios, including 2975 images for training and 500 images for validation. All images are at 1024 $\times$ 2048 resolution.
CamVid \cite{camvid} is also a real-world dataset for driving scenarios, which contains 367 training images, 101 validation images, and 233 test images with a resolution of 720 $\times$ 960. 
Cityscapes and CamVid are classical semantic segmentation datasets with multiple category annotations. 
When applying them to the road segmentation task, we set the label of the road to 1 and the other categories to 0.

\textit{2) Evaluation Metrics:}
For the KITTI-Road dataset, we evaluate the performance using six official metrics, namely maximum F1-measure (MaxF), average precision (AP), precision (PRE), recall (REC), false positive rate (FPR) and false negative rate (FNR).
Among them, MaxF is the main accuracy evaluation metric because it comprehensively considers precision and recall. 
It is worth noting that the metrics are computed in the Birds Eye View (BEV) for the KITTI-Road dataset as a common practice. 
We also evaluate the parameters, MACs, and inference time of our network. With $375 \times 1240$ as the input resolution, we compute the average time
of 1000 forwards on a single GPU as our inference time.
For Cityscapes and CamVid datasets, MaxF, PRE, REC, and mIoU (mean intersection over union) are employed to evaluate the performance in the image space.

\begin{table}[!t]
\caption{Model complexity comparison} \label{para}
\centering
\begin{tabular}{c|c|c}
\hline

Method & MACs [G] & Parameters [M] \\
 \hline
 \hline
 
FastRoadSeg\cite{gong2022fastroadseg} &18.323  &11.334  \\
\textbf{LFD-RoadSeg} & \textbf{8.392} & \textbf{0.936} \\
\hline
\end{tabular}
\end{table}

\subsection{Training Details}
\label{sec:train}

LFD-RoadSeg is implemented in PyTorch on Intel(R) Xeon(R) CPUs and is deployed on a single NVIDIA TITAN Xp GPU.  
The feature extractors of the two branches, namely part of the ResNet-18 \cite{resnet}, are loaded with the model parameters which are pre-trained on ImageNet \cite{imagenet} as initial weights. Other parameters of the LFD-RoadSeg are randomly initialized. Stochastic gradient descent (SGD) with a momentum of 0.9 and weight decay of $10^{-4}$ is used to optimize our network. The initial learning rate is set to 0.01 and the cosine annealing learning rate decay strategy is adopted in the training process. The final learning rate is set to $10^{-5}$. In order to prevent the model from overfitting, 
\huan{we apply some data augmentations on the training images such as random cropping, random horizontal flipping, random brightness adjusting with the range of [0.9, 1.1], and random scaling with the range of [0.5, 2.0].}
\huan{The OHEM loss threshold $\lambda_{b}$ is set to 0.7 in the experiment.} 
For KITTI-Road, we employ a two-step training scheme. First, we randomly crop out $320 \times 500$ patches from the original images as the inputs $I^{h}$ and second, the training is resumed by using the full images as $I^{h}$. The batch size for the first training step is set to 16 and for the second training step is set to 6. We train LFD-RoadSeg for 150 epochs in these two steps. 
For Cityscapes, $800 \times 800$ patches are randomly cropped out for training. The batch size is 4 and the maximum number of epochs is 250.
For CamVid, the input image size is $720 \times 960$. The batch size is 4 and the maximum number of epochs is 150.

\subsection{Final Results and Comparison with Prior Works}
\label{sec:result}
\begin{figure}[!t]
\centering
\includegraphics[width=1.0\linewidth]{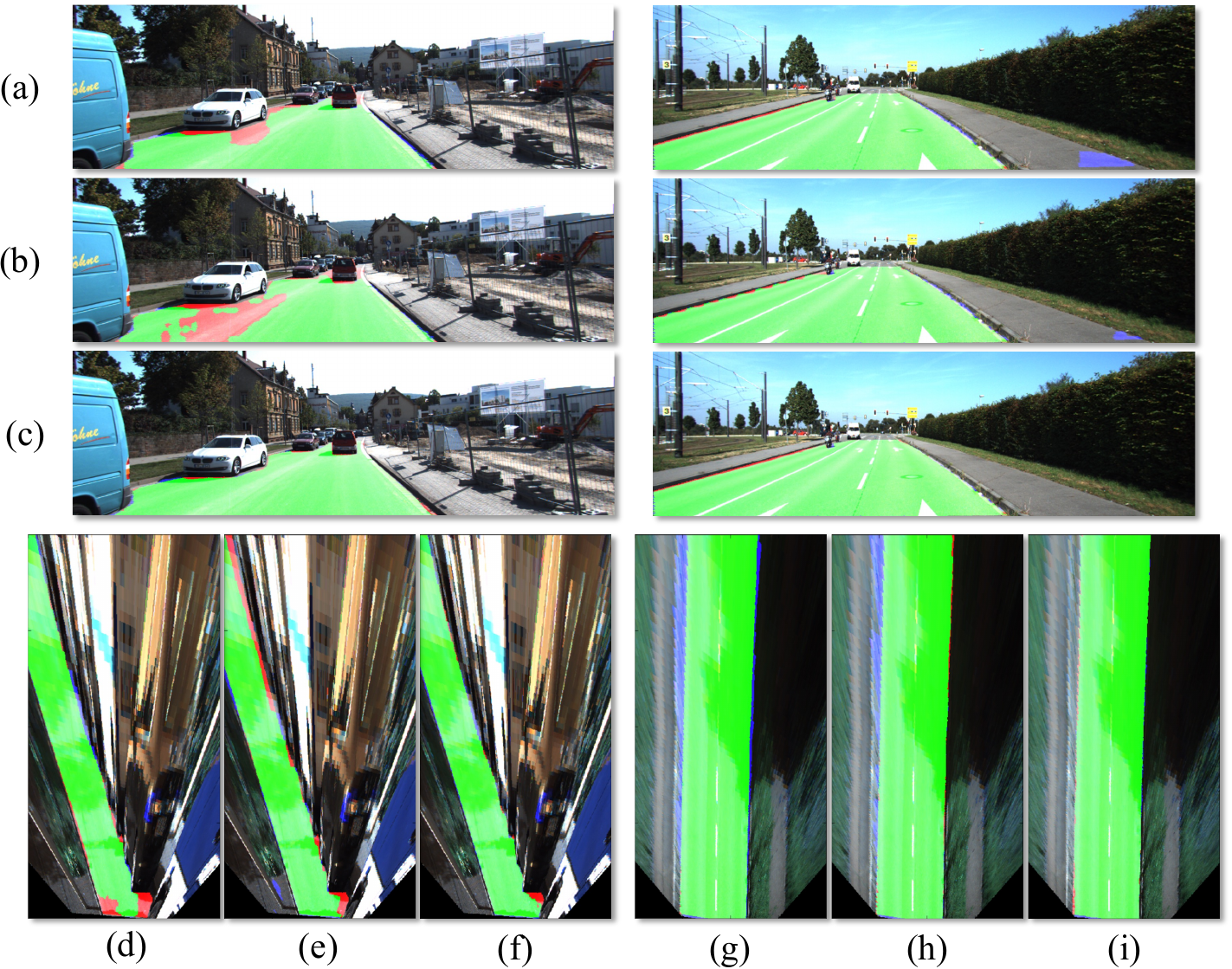}
\caption{Qualitative comparison with other real-time monocular road segmentation methods. (a)-(c) are in the camera's perspective view and (d)-(i) are in the birds eye view. (a) (d) (g): ALO-AVG-MM \cite{Reis:1}, (b) (e) (h): RoadNet-RT \cite{bai2020roadnet}, (c) (f) (i): LFD-RoadSeg. Red marks false negatives, blue marks false positives, and green marks true positives.}
\label{vis}
\end{figure}

\huan{\textbf{KITTI-Road dataset.}
Table \ref{leaderboard} reports the segmentation results and the corresponding time cost comparison between LFD-RoadSeg and the prior representative road segmentation works on the KITTI-Road leaderboard.
\xf{Note that, we cannot compare all methods on the same platform as most of them did not release the code.
Therefore, we use the ``Device" and ``Time" information provided by the official KITTI benchmark database and each paper to comprehensively evaluate the computational efficiency of each method.}
From Table \ref{leaderboard}, we can see that LFD-RoadSeg is the fastest and outperforms many monocular-based methods \cite{rbnet,garnett2017real,oeljeklaus2021integrated,teichmann2018multinet,fan2021learning,roadnet3}
in MaxF and achieves a high AP of 93.71\%. 
In addition, RBANet \cite{rbanet}
and SSLGAN \cite{han2018semisupervised} fail to meet real-time requirements. 
Compared with the current second-fastest method FastRoadSeg \cite{gong2022fastroadseg}, LFD-RoadSeg only reduces MaxF by 0.35\% (95.56\%-95.21\% = 0.35\%), but the speed is increased by 43.2\% (1-4.2/7.4 = 43.2\%).
Furthermore,
Table \ref{para} shows the complexity of FastRoadSeg \cite{gong2022fastroadseg} and LFD-RoadSeg.
LFD-RoadSeg has more than 12 times fewer parameters than FastRoadSeg and reduces the MACs  (multiply-accumulate operations) by 54.2\% (1-8.392/18.323 = 54.2\%).
Compared with RoadNet-RT \cite{bai2020roadnet}, which is also a bilateral network, LFD-RoadSeg increases the MaxF from 92.55\% to 95.21\%,
which is due to the more reasonable network depth and more efficient fusion process.
The detailed performance provided by the KITTI online test server is shown in Table \ref{LFD-RoadSeg}.
Fig. \ref{vis} illustrates the qualitative comparison with other real-time monocular methods, ALO-AVG-MM \cite{Reis:1} and RoadNet-RT \cite{bai2020roadnet}.
We show the camera's perspective view and birds eye view respectively.
Compared to other methods, LFD-RoadSeg reduces both the number of false positive and false negative pixels.
}

\huan{\textbf{Cityscapes dataset.}}
\xf{
Table \ref{city} presents the performance of several methods on the Cityscapes dataset with metrics such as MaxF, PRE, REC, and mIoU.
In terms of the MaxF metric, despite LFD-RoadSeg fails to surpass the best method, i.e., RBANet \cite{rbanet},
it outperforms conventional benchmark methods like FCN [27],
and it is only 0.47\% lower than FastRoadSeg \cite{gong2022fastroadseg} on MaxF.
Not that, the accuracy gap between our method and the best method lies in precision (3.59\% lower),
while our recall is not much different from that of the best method (only 0.33\% lower).
The reason is that the Cityscapes dataset contains a large number of specific categories of objects,
that is, ``things'' defined in panoptic segmentation.
Using low-level features to express these objects during training would slightly degrade the discriminative ability of the model.
}


\begin{table}[!t]
\scriptsize
\centering
\caption{Performance evaluation from KITTI online test server}
\label{LFD-RoadSeg}
\begin{tabular}{c|c|c|c|c|c|c}
\hline
Benchmark &  MaxF   & AP  & PRE  & REC  & FPR  & FNR \\
\hline
\hline
\tiny{UM\_ROAD} &94.58\% &93.42\% &95.20\% &93.98\% &2.16\% &6.02\% \\
\tiny{UMM\_ROAD} &96.59\% &95.40\% &96.29\% &96.90\% &4.11\% &3.10\%
\\
\tiny{UU\_ROAD} &93.49\% &92.19\% &93.46\% &93.52\% &2.13\% &6.48\%
\\
\tiny{URBAN\_ROAD} &95.21\% &93.71\% &95.35\% &95.08\% &2.56\% &4..92\%
\\
\hline
\end{tabular}
\end{table}

\begin{table}[!t]
\scriptsize
\centering
\caption{Road segmentation results on Cityscapes dataset\\
(\huan{``-" means it is not mentioned in the original paper,} \xf{$\dagger$ denotes the reproduced results by \cite{gong2022fastroadseg}.})}
\label{city}
\resizebox{1\linewidth}{!}{
\begin{tabular}{c| c c c c}
\hline
Methods &MaxF(\%)$\uparrow$  & PRE(\%)$\uparrow$ &  REC(\%)$\uparrow$  & mIoU(\%)$\uparrow$ \\
\hline
\hline

 
 

 Zohourian \etal\cite{Zohourian:1}  & 92.44 & 89.08 & 96.76  &-\\
  FCN \cite{fcn} & 94.68 &  93.69 &  95.70 &-\\
  $^{\dagger}$FCN \cite{fcn} & 94.75 &  93.65 & 95.77 &93.96 \\

  s-FCN-loc \cite{wang:3}  & 95.36 & 94.63 & 96.11  &-\\
  SegNet \cite{segnet} & 95.81 & 94.55 & 97.11  &-\\
  $^{\dagger}$SegNet \cite{segnet} & 95.92 &  94.73 &  96.78 & 94.33\\
 
 $^{\dagger}$U-Net\cite{unet} & 96.26 & 94.89& 97.19 & 95.21\\
  FastRoadSeg \cite{gong2022fastroadseg} &96.48 & 95.84 & 97.12 & 95.74\\

 $^{\dagger}$FASSD-Net \cite{rosas2021fassd} & 97.47  & 97.51 & 98.07 &96.34\\
 RBANet \cite{rbanet} & 98.00 & 97.87 & 98.13 &-\\
\hline
 \textbf{LFD-RoadSeg} &\textbf{96.01}  &\textbf{94.28}  &\textbf{97.80}  &\textbf{96.68} \\

\hline
\end{tabular}
}
\vspace{5pt}

\end{table}

\begin{table}[!t]
\scriptsize
\centering
\caption{Road segmentation results on CamVid dataset\\(\huan{``-" means it is not mentioned in the original paper}, \xf{$\dagger$ denotes the reproduced results by \cite{gong2022fastroadseg}.})}
\label{camvid}
\resizebox{1\linewidth}{!}{
\begin{tabular}{c| c c c c}
\hline
Methods & MaxF(\%)$\uparrow$  & PRE(\%)$\uparrow$ &  REC(\%)$\uparrow$  & mIoU(\%)$\uparrow$ \\
\hline
\hline
 RoadNet-RT \cite{bai2020roadnet}   & 92.98 &  94.70 &  91.91 & -\\
  SegNet \cite{segnet} & 93.95 &  93.07 &  94.86 & -\\
  $^{\dagger}$SegNet \cite{segnet} & 94.89 &  94.27 &  95.91 &94.13\\

  Yadav \etal\cite{Yadav:1} & 94.14 & 93.31 & 94.99  & -\\
 
 $^{\dagger}$U-Net\cite{unet} &96.45 & 95.73& 96.11 & 95.65\\
  RBANet \cite{rbanet} & 96.72 & 97.14 & 96.30 & -\\
 
  FastRoadSeg \cite{gong2022fastroadseg} & 97.02 & 96.79 & 97.24 & 96.11\\
 $^{\dagger}$FASSD-Net \cite{rosas2021fassd} & 97.38  & 97.69 & 98.81 & 97.64\\
\hline
  \textbf{LFD-RoadSeg} &\textbf{97.02}  &\textbf{96.80} &\textbf{97.25}  &\textbf{95.70} \\

 



\hline
\end{tabular}
}
\vspace{5pt}


\end{table}

\begin{table*}[!t]
\centering
\caption{Effectiveness of each component on KITTI-Road cross-validation.
(SDB: Spatial Detail Branch;
CSB: Context Semantic Branch.}

\begin{tabular}{ c | c c | c c c c | c | c c c | c}
\hline

 SDB & CSB &Aggregation Module &$\circled{c}$ &$\otimes$ &$\oplus$ & Selective Fusion & Classifier &MaxF(\%)$\uparrow$ & PRE(\%)$\uparrow$ &  REC(\%)$\uparrow$ &Parameters\\
 
\hline
\hline

 \Checkmark &    & &&&& & \Checkmark &91.13$\pm$0.18 &88.90$\pm$0.46 &93.49$\pm$0.40 &183106\\
 ~ & \Checkmark   & && &&& \Checkmark &94.69$\pm$0.28 &94.51$\pm$0.52 &94.88$\pm$0.43 &700098\\
   ~ &\Checkmark     &\Checkmark & &&&&\Checkmark  &94.88$\pm$0.17 &94.80$\pm$0.56 &94.96$\pm$0.43&736706\\
  \Checkmark & \Checkmark   && &&&\Checkmark &\Checkmark &95.76$\pm$0.25 &95.78$\pm$0.45&95.75$\pm$0.42&899459\\
  \hline
  \Checkmark &\Checkmark    &\Checkmark &\Checkmark && &&\Checkmark &95.96$\pm$0.21 &96.03$\pm$0.35&95.89$\pm$0.21&919170\\
  \Checkmark &\Checkmark    &\Checkmark& &\Checkmark && &\Checkmark &95.77$\pm$0.16 &96.14$\pm$0.30&95.40$\pm$0.26&902786\\
  \Checkmark &\Checkmark    &\Checkmark& &&\Checkmark & &\Checkmark &95.92$\pm$0.10 &96.21$\pm$0.30&95.62$\pm$0.30&902786\\
  \hline
    \Checkmark &\Checkmark    &\Checkmark&&& &\Checkmark &\Checkmark &\textbf{96.28$\pm$0.14} &\textbf{96.56$\pm$0.11}&\textbf{96.00$\pm$0.21}&936067\\
\hline
\end{tabular}
\label{ablation}
\end{table*}

\huan{\textbf{CamVid dataset.}}
\xf{Table VII provides the quantitative comparisons on the CamVid dataset.
Notably, our method achieves an impressive MaxF score of 97.02\%, the same as FastRoadSeg \cite{gong2022fastroadseg} and even outperforming FastRoadSeg \cite{gong2022fastroadseg} in precision and recall.
Compared to RBANet \cite{rbanet}, our method achieves 0.3\% higher MaxF than RBANet \cite{rbanet} which is a non-lightweight method.
Compared to RoadNet-RT \cite{bai2020roadnet} that also utilizes a bilateral architecture, our method obtains significantly higher accuracy in MaxF (4.04\% higher), PRE (5.20\% higher), and REC (1.04\% higher) metrics.
In contrast, RoadNet-RT \cite{bai2020roadnet} requires 9 ms for inference at a resolution of $280 \times 960$ on a GTX 1080 GPU,
whereas LFD-RoadSeg achieves inference in just 4.2 ms at $375 \times 1240$ resolution on a TITAN Xp GPU,
showcasing superior speed performance.

Overall, the performance of our method is close to that of the State-of-the-art method on the KITTI-Road and CamVid datasets.
This is due to the fact that in urban street scenes, low-level features are sufficient to represent most road areas.
}

\subsection{Ablation Study of Each Component}
\label{sec:ablation}
We further examine the effectiveness of each component in the proposed LFD-RoadSeg. From Table \ref{ablation}, we can see that the performance of LFD-RoadSeg is better than using the spatial detail branch only and the context semantic branch only.
The reason is that the low-level features are not effective in distinguishing different long-range areas due to their limited local receptive field,
while the context feature lacks spatial detail due to the low resolution.
As shown in Table \ref{ablation},
when using the context semantic branch only,
the aggregation module helps to improve the MaxF from 94.69\% to 94.88\%.
When using both branches,
the aggregation module helps to improve the MaxF from 95.76\% to 96.28\%.
The two comparisons indicate that the aggregation module is suitable for capturing the context information with a large receptive field.
Furthermore,
we verify the effectiveness of selection fusion,
and compare it with three fusion processes,
namely, direct concatenation, direct element-wise product and direct element-wise addition.
As we can see from the last four rows of Table \ref{ablation}, the selection fusion achieves the best MaxF, PRE, and REC with only a few additional parameters.
Finally, when we train the network with all components, the proposed LFD-RoadSeg achieves the best performance on the KITTI-Road dataset.

\begin{table}[!t]
\caption{Performance comparison of the context semantic branch with different input image sizes on KITTI-Road cross-validation} 
\label{size}
\centering
\begin{tabular}{c c | c | c c}
\hline
Height &Width &MaxF(\%)$\uparrow$ & MACs [G] &Time (ms)  \\
 \hline
 \hline
1/2 &1/2 &96.32$\pm$0.20 &9.598 &4.6\\
\textbf{1/2} &\textbf{1/4} &\textbf{96.28$\pm$0.14} &\textbf{8.392} &\textbf{4.2}\\
1/4 &1/2 &95.86$\pm$0.18 &8.400 &4.2\\
1/4 &1/4 &96.01$\pm$0.23 &7.792 &4.1\\

\hline
\end{tabular}
\end{table}

\begin{table}[!t]
\caption{\huan{Performance comparison of the context semantic branch with different input image sizes on Cityscapes and CamVid datasets}} 
\label{size2}
\centering
\begin{tabular}{c | c | c | c | c | c}
\hline
\huan{Datasets} &\huan{Height} &\huan{Width} &\huan{MaxF(\%)$\uparrow$} &\huan{PRE(\%)$\uparrow$} &\huan{REC(\%)$\uparrow$}  \\
 \hline
 \hline
\multirow{4}{*}{\huan{Cityscapes}}
&\huan{1/2} &\huan{1/2} &\huan{95.83} &\huan{94.23} &\huan{97.50}\\
&\huan{\textbf{1/2}} &\huan{\textbf{1/4}} &\huan{\textbf{96.01}} &\huan{\textbf{94.28}} &\huan{\textbf{97.80}}\\
&\huan{1/4} &\huan{1/2} &\huan{95.58} &\huan{94.12} &\huan{97.08}\\
&\huan{1/4} &\huan{1/4} &\huan{95.87} &\huan{94.53} &\huan{97.26}\\
\hline
\multirow{4}{*}{\huan{CamVid}}
&\huan{1/2} &\huan{1/2} &\huan{96.20} &\huan{95.75} &\huan{96.65}\\
&\huan{\textbf{1/2}} &\huan{\textbf{1/4}} &\huan{\textbf{97.02}} &\huan{\textbf{96.80}} &\huan{\textbf{97.25}}\\
&\huan{1/4} &\huan{1/2} &\huan{95.80} &\huan{94.82} &\huan{96.80}\\
&\huan{1/4} &\huan{1/4} &\huan{96.41} &\huan{96.03} &\huan{96.79}\\

\hline
\end{tabular}
\end{table}

\subsection{Discussion on Input Resolution}
\label{sec:discussion}
The input image $I^l$ for the context semantic branch is obtained by asymmetrically downsampling the original image $I^h$,
which has two goals.
The first goal is to reduce the computational burden as much as possible and speed up the training and testing processes.
The second goal is to gain a larger horizontal receptive field so that LFD-RoadSeg captures longer-range dependencies in the horizontal direction.
The results using different aspect ratios are shown in Table \ref{size}.
Compared to our asymmetric input size setting,
when both height and width are reduced to half of the original size,
the MaxF only increases from 96.28\% to 96.32\% but at the cost of 0.4 ms longer inference time.
And when both height and width are shrunk by a quarter,
the model increases the MaxF from 96.01\% to 96.28\% and only sacrifices 0.1 ms in inference time.
The two comparisons illustrate that our experimental setting has a better trade-off than others.
Furthermore,
when inverting the downsampling rate of height and width,
that is, the network has a larger vertical receptive field,
the MaxF drops dramatically and is even worse than the model using a smaller resolution.

\begin{figure}[!t]
\centering
\includegraphics[width=1.0\linewidth]{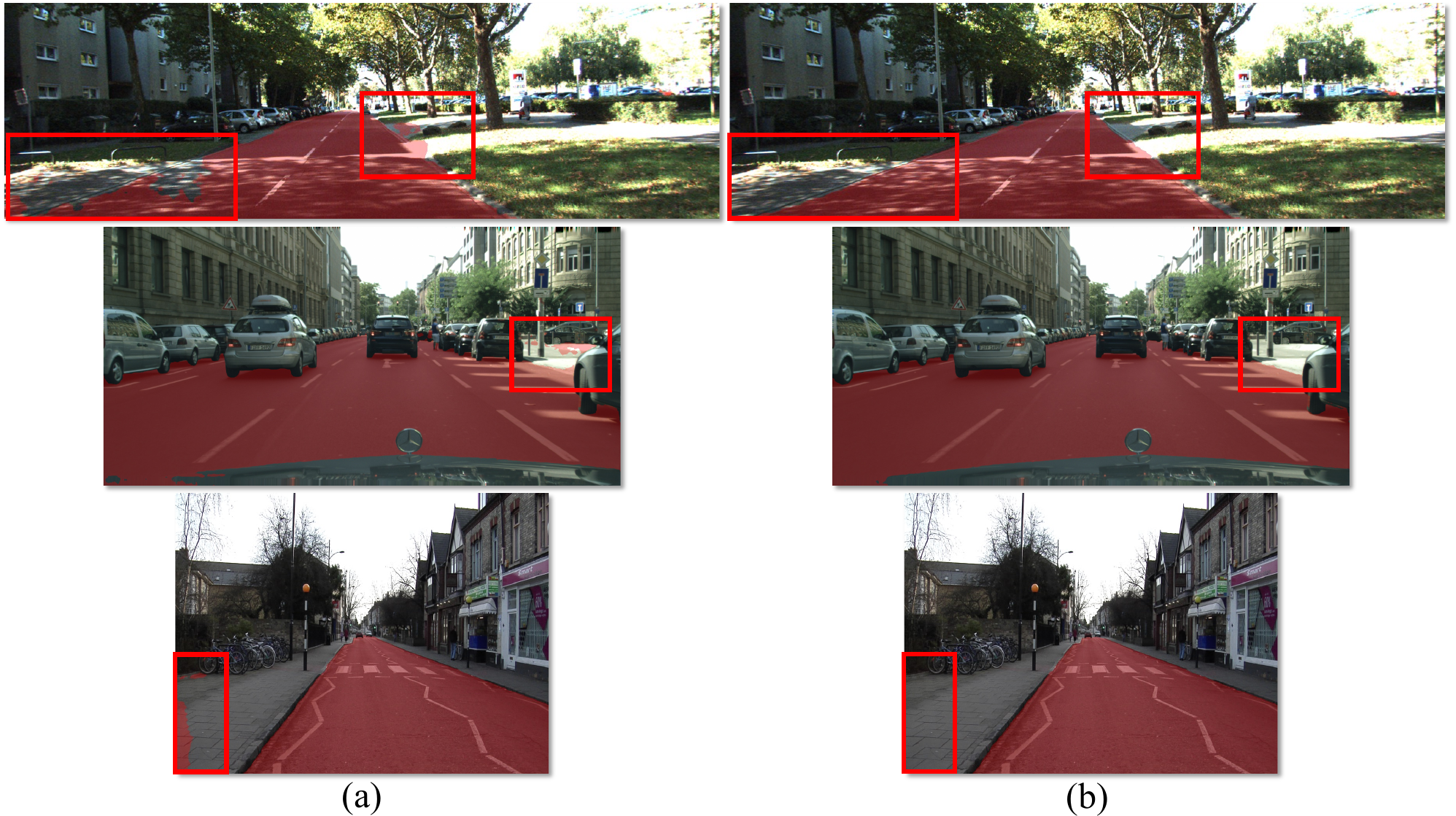}
\caption{\huan{The visualization of the road prediction result, where the area covered by red is the road. (a): The input image size of the context semantic branch is 1/4 the height of the original image and 1/2 the width of the original image. (b): The input image size of the context semantic branch is 1/2 the height of the original image and 1/4 the width of the original image.}}
\label{2442}
\end{figure}

\xf{
We then discuss the asymmetric downsampling on the Cityscapes and CamVid datasets in Table \ref{size2}.
Our experimental setting (the bolded row in Table \ref{size2}) yields the highest MaxF scores,
while the opposite setting results in the lowest MaxF scores.
The MaxF gaps between the two reach 0.43\% on Cityscapes and 1.22\% on CamVid.
It verifies that the significance of horizontal receptive fields to deep networks is not limited to the KITTI-Road dataset,
but is ubiquitous in various street scenes.
Fig. \ref{2442} further displays the visual comparisons between our setting and the opposite setting.
The first to third rows are from the KITTI-Road, Cityscapes, and CamVid datasets.
Observably,
the model with the opposite setting obtains many false positives on the texture-less area outside the road as it has a narrow receptive field,
which is highlighted by the red rectangular boxes.
However, our model with the proposed asymmetric downsampling setting eliminates most false positives.
}


\huan{
\subsection{Results and Comparison for Different Backbones}
\label{sec:backbones}

Table \ref{backbone} provides a performance comparison of \xf{lightweight backbones with various structures} on the KITTI-Road dataset,
including ResNet-18 \cite{resnet},
MobileNetV2 \cite{sandler2018mobilenetv2} and MobileViT-XXS \cite{mehta2021mobilevit}.
We measure them based on MaxF, AP, parameters and inference time.

Specifically, for ours (ResNet-18), the spatial detail branch utilizes the first stage of ResNet-18 as its backbone, while the context semantic branch employs the first two stages of ResNet-18 as its backbone.
For ours (MobileNetV2) and ours (MobileViT-XXS), due to the lack of official stages for MobileNetV2 and MobileViT, we use layers that extract features at 1/4 the original resolution as the spatial detail branch, and layers that extract features at 1/8 the original resolution as the context semantic branch.

\begin{table}[!t]
\scriptsize
\caption{\huan{Performance comparison of different backbones on KITTI-Road dataset}} 
\label{backbone}
\centering
\begin{tabular}{ c | c | c | c | c}
\hline
\huan{Methods} &\huan{MaxF(\%)$\uparrow$} &\huan{AP(\%)$\uparrow$} &\huan{Params} &\huan{Time (ms)}\\
 \hline
 \hline

\huan{ResNet-18} &\huan{95.08} &\huan{93.74} &\huan{11,177,538} &\huan{4.58}\\
\huan{Ours (ResNet-18)} &\huan{96.39} &\huan{94.19} &\huan{936,067} &\huan{4.24}\\
\hline
\huan{MobileNetV2} &\huan{95.55} &\huan{93.90} &\huan{1,884,162} &\huan{9.05}\\
\huan{Ours (MobileNetV2)} &\huan{96.06} &\huan{94.16} &\huan{161,395} &\huan{7.49}\\
\hline

\huan{MobileViT-XXS} &\huan{95.15} &\huan{93.85} &\huan{947,554} &\huan{22.75}\\
\huan{Ours (MobileViT-XXS)} &\huan{95.98} &\huan{94.11} &\huan{253,779} &\huan{8.20} \\

\hline
\end{tabular}
\end{table}

From Table \ref{backbone}, observably, ours (ResNet-18) achieves the highest MaxF, reaching 96.39\%, and the highest AP, reaching 94.19\%. Compared with the original backbone network (i.e., ResNet-18, MobileNetV2, and MobileViT-XXS), all variants equipped with the proposed structure achieve higher accuracy (MaxF improvement of 1.31\%, 0.51\%, 0.83\%, respectively), lower parameters (reduced by 99.99\%, 91.43\%, 73.22\% respectively) and less inference time (reduced by 7.42\%, 17.24\%, 63.95\% respectively). This experiment proves that the proposed structure is not only suitable for classic residual networks, but also advanced transformer networks. Even on the transformer model, our structure contributes to a significant speed improvement.

\subsection{Deployment and Speed Comparison}

Following the works of \cite{gong2022fastroadseg}\cite{chen2023cooperative}\cite{chen2023edge}, we deployed the proposed model on the Jetson TX2 \cite{tx2}. The Jetson TX2 is equipped with a quad-core ARM A57 processor, a dual-core Denver2 processor, a 256-core NVIDIA Pascal™ architecture GPU, and 8GB of 128-bit LPDDR4 memory. This configuration makes it suitable for use in robots, drones, and other intelligent edge devices. 
During the inference process, the input image is resized to $187\times620$, aligning with FastRoadSeg \cite{gong2022fastroadseg}. Table \ref{deploy} provides details on the speed and power consumption in both maximum processing efficiency (Max-N) and maximum energy efficiency (Max-Q) modes on the Jetson TX2 and offers a comparison with FastRoadSeg \cite{gong2022fastroadseg} on the same platform. 
Note that all methods do not use any acceleration techniques. Observably, LFD-RoadSeg is 74\% faster than FastRoadSeg with 2W lower power consumption. When using less than 7W of power, its speed is still 35\% faster than that of FastRoadSeg \cite{gong2022fastroadseg}.
}



\begin{table}[!t]
\caption{\huan{Deployment on Jetson TX2\cite{tx2}}} 
\label{deploy}
\centering
\begin{tabular}{c | c | c | c }
\hline
\huan{Method} &\huan{Runtime} &\huan{Frame Rate} &\huan{Power Consumption}  \\
 \hline
 \hline
\huan{FastRoadSeg\cite{gong2022fastroadseg}} &\huan{32.2 ms} &\huan{31 FPS} &\huan{14.8 W(3.1 W idle)} \\
\huan{LFD-RoadSeg (Max-N)} &\huan{18.5 ms} &\huan{54 FPS} &\huan{12.8 W(3.1 W idle)} \\
\huan{LFD-RoadSeg (Max-Q)} &\huan{14.4 ms} &\huan{42 FPS} &\huan{6.9 W(2.3 W idle)} \\

\hline
\end{tabular}
\end{table}

\section{Conclusion}
\label{conclusion}
\huan{

In this study,
considering that roads are part of the environmental background rather than specific objects,
we propose a Low-level Feature Dominated Road Segmentation network (LFD-RoadSeg) \xf{to achieve accurate and efficient road segmentation}.
It follows a bilateral structure.
The spatial detail branch \xf{extracts low-level road representation}.
\xf{The context semantic branch quickly captures the context having large horizontal receptive fields with the help of asymmetric downsampling and lightweight aggregation modules.}
In addition,
\xf{the selective fusion module leverages the context to suppress the non-road response in the low-level feature.}
Comprehensive experiments on three datasets indicate that our method achieves similar accuracy as mainstream methods but at a speed of 238 FPS on a single TITAN Xp, 54 FPS on a Jetson TX2 and costs the parameter amount of only 936k.

\section{Limitations and Future Work}
While our research has made significant strides in real-time road segmentation, there are several limitations to acknowledge.
Our model performs well under typical road conditions.
However, it may face challenges in extreme weather conditions,
such as heavy rain, snow, or fog.
Future work should focus on enhancing the model's robustness under adverse conditions. 
\xf{In addition, due to using low-level features, our model cannot be generalized to the scenes totally different from the training set,
such as training models on street scenes and testing models on off-road scenes.}
In the future, it will be possible to enhance the model's robustness and generalization by utilizing techniques such as training on multiple datasets, data augmentation, and transfer learning, thereby improving its \xf{practicability}.
}

\bibliography{FRSv7.bib}
\bibliographystyle{IEEEtran}

\vfill

\end{document}